# FACIAL EXPRESSION RECOGNITION IN THE WILD USING RICH DEEP FEATURES

*Abubakrelsedik Karali, Ahmad Bassiouny and Motaz El-Saban*

Microsoft Advanced Technology labs, Microsoft Technology and Research, Cairo, Egypt

## ABSTRACT

*Facial Expression Recognition is an active area of research in computer vision with a wide range of applications. Several approaches have been developed to solve this problem for different benchmark datasets. However, Facial Expression Recognition in the wild remains an area where much work is still needed to serve real-world applications. To this end, in this paper we present a novel approach towards facial expression recognition. We fuse rich deep features with domain knowledge through encoding discriminant facial patches. We conduct experiments on two of the most popular benchmark datasets; CK and TFE. Moreover, we present a novel dataset that, unlike its precedents, consists of natural - not acted - expression images. Experimental results show that our approach achieves state-of-the-art results over standard benchmarks and our own dataset.*

## 2. INTRODUCTION

Automatic recognition of facial expressions, as a branch of computer vision, has gained increasing attention due to its wide range of applications, such as human-computer interaction, social analytics, medical treatment and intelligent robot systems. In the literature, each facial movement can parameterized using two major strategies: Facial Action Coding System (FACS)[1] and Facial Animation Parameters (FAPs)[2]. This parameterization is used to describe primary expressions Facial Expressions (FEs) [3]. While the FACS strategy encodes the movements in action units that are based on facial muscle movements, FAPs is based on 84 predefined facial feature point movements, and FEs detects emotions categorized into seven major expressions: surprise, fear, disgust, contempt, happiness, sadness, anger and neutral. The research done in facial expressions recognition can be categorized into two major categories[3]: Dynamic approaches; which deal with recognizing facial expressions from a video stream or image sequence and static approaches; which deal with recognizing the emotions from still images. In this paper, we deal with recognizing facial expressions from static images. For the rest of this paper wherever "Facial expressions recognition" is used, it refers to recognition from static images. Facial expressions extracted from videos, however, is a possible extension of this work.

A typical static facial expression recognition approach could be divided into four main components[3] face detection, face alignment, feature extraction and finally classification. As for the step of feature extraction of expression, most of the recent approaches presented in the literature design a handcrafted feature extraction method or uses a combination of many features [4]. In contrast, this paper is based on rich deep features extracted using Deep Convolutional Neural Networks[1] (CNNs)[5]. CNNs have recently become a usual step towards attempting various vision problems [6] replacing the hand-crafted features for specific problems with learned ones based on readily available large pool of images. In [6] CNNs achieved very comparable results to those from the state-of-the-art in and was successfully applied for scene classification [8] and object recognition[9].

One of the essential aspects in developing a facial expression recognition system is the database that will be used for testing the system. Common databases used in literature, such as CK+[10] TFEID[11] and MMI[12], are collected using actors, where each actor performs the actions in an indoor experimental setup. However, the acting nature in those datasets sentences them to being only a subset from natural daily-life actions. In this paper, besides measuring our performance against standard datasets as a baseline, we also present a novel dataset contains natural expressions images collected from Bing image search engine and is available online: Facial Expressions in the Wild. In summary, this paper makes the following contributions: (i) Development of rich deep features that exceeds the state-for-the-art in facial expression recognition; (ii) Utilizing domain knowledge by extracting deep features from discriminant facial patches; and (iii) Presenting a novel dataset that deals with natural expression images that is aimed at encouraging the trend in facial expressions research towards studying more realistic cases. The paper is structured as follows: In section 2 we present related work. Section 3 describes our approach. Section 4 and 5 describes the datasets we use and the conducted experiments respectively. Finally our conclusion and future work suggestions are presented in section 6.

## 3. RELATED WORK

Researchers have proposed multiple methods and systems for automatic facial expression recognition. For face detection, many approaches exist in the literature including the popular approach by Viola and Jones [13] used for real-time detection using Haar features. Hoai et al. [14]developed max-margin early event detectors based on a structured output support vector machine (SVM). The detectors can recognize partial human motion events including facial expressions, enabling early detection. El-Bakry[15] presented a fast method for face detection based on principal component analysis (PCA) by performing cross-correlation between eigenvectors and the input images in the frequency domain.

---

[1] Deep Neural Networks (DNNs) have been used in [37] however dealing with videos and is based on a multimodal approach using audio as well.

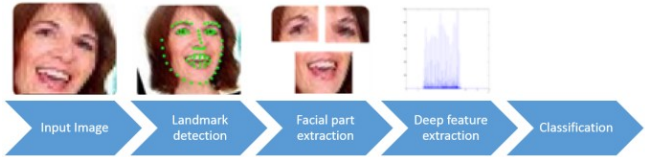

Figure 1 The proposed approach in steps: the image is input to the landmark detector, then facial parts are extracted followed by deep features extraction and finally classification

As for representation, there exist mainly two kinds of features that are widely adopted for facial representation: geometric features and appearance features [16]. In [17] the detailed geometric positions of 34 fiducially points on a facial image are adopted for a facial representation. Extracting geometric features still needs facial feature detection in order to achieve precision and reliability, which is rather hard to achieve in real-time practical scenarios. In addition, geometric features would not effectively encode changes in texture, like wrinkles and furrows, which are crucial in modeling facial expressions. In contrast to geometric features, appearance features are capable of reflecting texture variations. Popular appearance features techniques include Gabor wavelets representation [18], Eigenfaces[19],Fisherfaces[19] as well as raw pixel values of facial images[20]. In addition, local binary patterns (LBP) [21], as a face descriptor, has received increasing interest during the past few years. So far, LBP is widely employed as appearance features for facial expression classification [22] owing to its tolerance against illumination variations and its low computation cost.

Lastly, classification/recognition methods have been adopted for frame-based facial expression recognition systems, including SVM[23], nearest neighbor (NN) or K-nearest neighbor (KNN) [24] and artificial neural network (ANN) [25]. On the other hand classification methods for sequence-based facial expression classification typically include hidden Markov models (HMM) [26] and dynamic Bayesian networks [27]. In this paper we are mainly concerned with the representation part for feature encoding and hence, in principle, many different classifiers can be combined with our work.

## 3. APPROACH

In this section we will describe our approach towards automatic facial expression recognition. We use a 3D head pose estimation face detection method to cope with head rotation[28]. Then, we extract discriminative patches for each face based on facial landmarks locations. We then extract rich deep features from the face and its discriminative patches. These features are then pre-processed before being used as input for classification.

### 3.1. Face detection

We leverage the proposed framework [28] to detect facial landmarks and 3D head pose. The framework has the advantage of handling large facial pose variations. It is based on training mixtures of trees with a shared pool of parts, where a part is the patch around a facial landmark. Finally, all model parameters, including part templates, modes of elastic deformation, and view-based topology, are discriminatively trained by a max-margin framework. Before training initialization, the facial anchor point is selected using a group sparse learning method. Then, a two-level cascaded deformable shape model is used to search for global optimal positions of the landmarks. After detecting the face, including the landmarks and the pose parameters, we perform in-plane rotation to vertically align the faces. Then, we extract predefined discriminative patches: the two eyes and the mouth. These are later used as input in the feature extraction step.

### 3.2. Deep features

Recently pre-trained CNN models over very large datasets have been rather successful in various computer vision problems and even managed to break state-of-the-art results in more than one occasion [8, 9]. For our feature extraction, this fact is exploited and an Imagenet-trained CNN is used to extract features representing images, with no need for a hand-tuned feature extraction. Deep features are extracted for each discriminative patch and for the detected face. Due to the high resulting dimensionality, we use Laplacian Eigenmaps [29] for dimensionality reduction and the number of new basis is determined using Maximum Likelihood estimation [30].

### 3.3. Classification

We have experimented with multiple classification techniques such as random forests, neural networks and logistic model trees [31] (LMTs). As will be shown in the experiments, LMTs perform best. LMTs is a supervised learning algorithm that combines

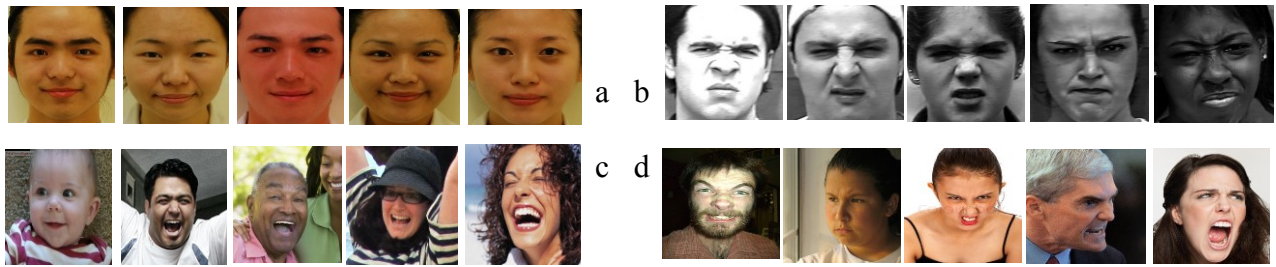

Figure 2 Example images from different datasets a) happy expression from TFE, b) disgust expression from CK c) happy expression from FEW and d) angry expression from FEW. The images show a clear difference between acted and natural expressions

logistic regression and tree induction. These two techniques have complementary advantages and disadvantages. Linear logistic regression performs a least square fit of the data to a numeric class. This results in low variance but may be prone to high bias.

Induction trees on the other hand can create models that capture nonlinear patterns in the data, which result in low bias but potentially high variance. Our logistic model tree is built by fitting logistic regression models on a tree node using training samples, then refining the model by adding more logistic regression models. These models are trained using subsets of the training data on the node's children. As a result, the final model at a leaf consists of a committee of linear regression models that have been trained on increasingly smaller subsets of the data. The tree is then pruned using the Classification and Regression Trees (CART) pruning algorithm [32].

## 4. EXPERIMENTAL SETUP

In this section we will first describe the used datasets and then we will present implementation details and the experimental setup.

### 4.1. Datasets:

**Cohn-Kanade Database (CK+)[10]:** The CK+ dataset contains 500 image sequences from 100 actors. The age of the samples ranges from 18 to 30 years with 65% female. It also contains 15% African Americans and 3% of Asians and Latinos. It contains seven facial expressions namely: surprise, fear, disgust, contempt, happiness, sadness and anger. Each actor performs the expressions once and the dataset is recoded with a clear background with only frontal head poses.

**Taiwanese Facial Expression Image Database (TFE)[11]:** The TFE dataset consists of images of 40 Taiwanese actors' faces showing eight different classes of expressions, namely: surprise, fear, disgust, contempt, happiness, sadness, anger and neutral expressions. The image provided is the peak frame that represents the action and is recoded with a clear back ground with frontal head poses.

**Facial Expressions Recognition in Wild (FEW):** We also collected a novel dataset: the FEW dataset consisting of 1408 natural images containing 6 facial expressions, namely: surprise, fear, disgust, happiness, sadness and anger. The images are collected from Bing Image search engine and then manually labeled. Examples from the datasets are shown in Figure 2 showing a clear difference between acted and natural expressions.

### 4.2. Methods

Throughout our experiments, we used the Caffe[33] open-source implementation for its wide usage across the research community, fast GPU implementation, and ease of comparability with other results produced using the same toolkit. Also, we used the pre-trained networks configuration provided by Caffe over the ImageNet challenge. Our network configuration consists of 8 layers: 5 convolutional, and 3 fully connected layers, with a final soft-max layer for output. Exact training steps are detailed in[7].

Our goal is to extract the pre-last layer activations; a 4096D feature vector. The input to the network is an array of 227x227 RGB image values. We conducted various experiments on best pre-processing practices for an image. Best results were obtained by first re-sizing an image to 256x256, then taking ten 227x227 crops: the four corners and the center, and the horizontal flipped image of each.

Then, we average the 4096D activations for each of the 10 crops to obtain the representative 4096D feature vector representing the image. Finally, we do L2 normalization. For each input Image we extract the deep features for the face and facial parts. Due to high resulting dimensionality, we use Laplacian Eigenmaps [29] for dimensionality reduction implemented in the dimensionality reduction toolbox [38]. The number of new basis is determined using Maximum Likelihood Estimation [30]. To reduce memory footprint, the dimensionality reduction step is done for all patches combined across training data and finally the resulting feature vectors for each image patches are concatenated to form a single feature vector representing the image. This feature vector is later used as input for classification.

## 5. RESULTS AND DISCUSSION

In the literature, due to the limited size of the benchmark datasets, several experiments [10,12] were conducted using three or more peak frames selected from each expression sequence to represent the action. Then, cross validation is applied for evaluation. However, such an experimental setup would lead the approaches to overfit on actor faces in the dataset. This leads to a significant improvement over these datasets [34]compared to poor performance for cross dataset evaluation [34] and even fails to recognize real cases, as shown later. In order to avoid the same problem and to have a real insight concerning our approach, we used only single peak frame for each expression per actor. Then, for all our experiments, 10 folds cross validation is applied. The evaluation is done using the average recognition accuracy and confusion matrices are presented whenever needed. Having the goal of improving on real cases, most of the analytical steps are tested over the FEW dataset. In addition, we compare our method to other state-of-the-art techniques over CK+, TFE and FEW datasets.

### 5.1. Landmark-based detection evaluation

We first compare results of our novel landmarks-based face detector (LM) to the most widely used Viola and Johns (VJ) detector. In this comparison we extract the facial parts and use raw features [20], LBP [34], and deep features for parts extracted using both VJ and LM. The training was done using random forests[35] with fixed parameters for all experiments. As per the presented results in Table 1, a marginal improvement is achieved on the CK+ and TFE benchmarks dataset, while significant improvement is achieved on the FEW datasets. We account this behavior by the fact that the FEW dataset contains rotated in-

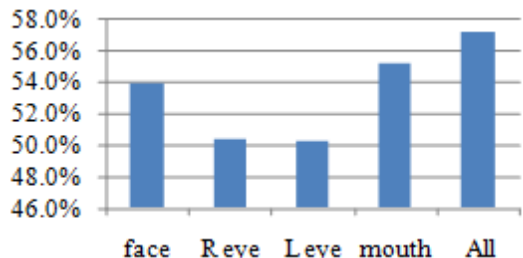

Figure 3 Recognition accuracy comparisons between individual facial parts features and features combination on plane faces. Unlike VJ, LM adjusts for these using pose-estimation calculations. The fact that both CK+ and TFE are only frontal and vertically oriented face solidifies our reasoning.

Table 1 Recognition accuracy comparison between LM based and VJ based facial expression recognition system using LBP [34], raw [20] and deep features.

| Face Detector | VJ | LM | VJ | LM | VJ | LM |
|---|---|---|---|---|---|---|
| Features | Raw | Raw | LBP | LBP | Deep | Deep |
| FEW | 50.9 | 53.6 | 44.8 | 50.4 | 52.3 | 59.1 |
| CK+ | 71.1 | 71.1 | 76.4 | 77.7 | 73.3 | 73.3 |
| TFE | 67.3 | 68.8 | 68.1 | 70.1 | 69.1 | 72.4 |

### 5.2. Face parts contribution

We evaluate the performance of each facial part individually on the FEW dataset before performing dimensionality reduction. Deep features are extracted, as shown in section 3.2, for each of the facial parts and for the whole face. Then we used random forests [35] for training. Recognition accuracy results presented in Figure 3 suggest that using multiple face parts achieves better results compared to the full face alone.

### 5.3. Dimensionality reduction and classification

For the purpose of studying feature dimensionality effect, we performed an experiment using extracted deep features from facial parts and another experiment using patches without dimensionality reduction. In this case, we performed training using random forests. Without Laplacian Eigenmaps dimensionality reduction with 52 dimensions, we achieve 57.2% recognition accuracy however, after reduction, accuracy improved by 2% to reach 59.1%. The accuracy increase can be explained by the fact that fewer features can lead to better generalization performance on unseen samples by avoiding overfitting.

We also evaluated several classification techniques: Random forest, back propagation neural networks and LMTs resulting in recognition accuracies of 59.1%, 62.2% and 64.1% respectively. While Neural Networks perform better than random forest, LMTs outperform all these techniques and score the best. We account this for the fact that the size of the training set is relatively small. LMTs are a special kind of the decision trees that employ a logistic regression function at the leaves of the tree, enabling it to perform better if the size of the training set is relatively small. Also, while building a tree, feature selection is used at every node yielding for better recognition accuracy. Table 2, 3 and 4 show confusion matrices of the LMTs on the FEW, CK+ and TFE datasets respectively with expressions abbreviated as follows: angry (An), contempt (Cn), disgust (Ds), fear (Fr), happy (Hp), sadness (Sd) and surprise (Sp).

Table 2 Confusion matrix for the FEW dataset using deep features with LMT for classification

|    | An   | Ds   | Fr   | Hp   | Sd   | Sp   |
|----|------|------|------|------|------|------|
| An | 30.9 | 9.8  | 8.9  | 29.3 | 15.4 | 5.7  |
| Ds | 7.3  | 40.6 | 2.4  | 30.3 | 15.2 | 4.2  |
| Fr | 2.1  | 5.3  | 52.4 | 15.3 | 6.9  | 18.0 |
| Hp | 1.3  | 3.3  | 1.8  | 86.7 | 3.8  | 3.1  |
| Sd | 2.5  | 12.7 | 2.0  | 16.2 | 60.4 | 6.1  |
| Sp | 2.7  | 2.7  | 14.2 | 15.8 | 8.2  | 56.3 |

Table 3 Confusion matrix for the CK+ dataset using deep features with LMT for classification

|    | An   | Cn   | Ds   | Fr   | Hp   | Sd   | Sp   |
|----|------|------|------|------|------|------|------|
| An | 83.3 | 4.2  | 6.3  | 2.1  | 2.1  | 2.1  | 0.0  |
| Cn | 14.3 | 66.7 | 0.0  | 4.8  | 0.0  | 9.5  | 4.8  |
| Ds | 10.3 | 1.7  | 82.8 | 0.0  | 0.0  | 3.4  | 1.7  |
| Fr | 0.0  | 0.0  | 16.7 | 50.0 | 16.7 | 16.7 | 0.0  |
| Hp | 1.2  | 0.0  | 6.0  | 11.9 | 79.8 | 1.2  | 0.0  |
| Sd | 31.3 | 6.3  | 6.3  | 18.8 | 0.0  | 37.5 | 0.0  |
| Sp | 0.0  | 0.0  | 1.1  | 3.4  | 0.0  | 4.5  | 90.9 |

Table 4 Confusion matrix for the TFE using deep features with LMT for classification

|    | An   | Cn   | Ds   | Fr   | Sd   | Hp   | Sp   |
|----|------|------|------|------|------|------|------|
| An | 78.4 | 3.9  | 2.0  | 3.9  | 5.9  | 2.0  | 3.9  |
| Cn | 7.4  | 63.0 | 7.4  | 14.8 | 3.7  | 3.7  | 0.0  |
| Ds | 5.6  | 5.6  | 83.3 | 0.0  | 0.0  | 2.8  | 2.8  |
| Fr | 11.4 | 2.3  | 9.1  | 65.9 | 6.8  | 4.5  | 0.0  |
| Sd | 0.0  | 9.5  | 7.1  | 4.8  | 76.2 | 2.4  | 0.0  |
| Hp | 5.9  | 8.8  | 2.9  | 0.0  | 0.0  | 82.4 | 0.0  |
| Sp | 8.1  | 0.0  | 2.7  | 5.4  | 0.0  | 0.0  | 83.8 |

We finally compare our approach to the state of the arts in the facial expressions recognition from static images. To the best of our knowledge, the state of the art was achieved by Shan et al.[34] and Přinosil et al. [20]. In Figure 4 we compare their approaches against ours using the experimental setups described in the beginning of this section.

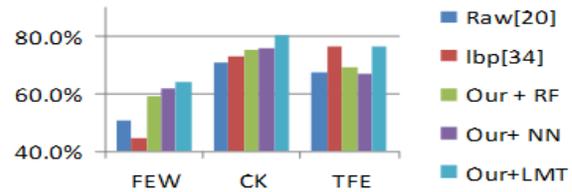

Figure 4 Accuracy results of our approach (with different classifiers) against the state-of-the-art approaches

## 6. CONCLUSION AND FUTURE WORK

We presented a novel approach towards facial expression recognition. We introduced the use of deep features alongside with domain knowledge for discriminative facial parts representation and conducted experiments on widely used benchmark datasets; CK+ and TFE. Moreover, we presented a novel dataset that consists of natural expression images. Experimental results showed that our approach outperforms the state-of-the-art over both standard benchmarks and our dataset. Extension to this work includes using large datasets to re-train the CNN. Spatial pyramids that are commonly used to improve the scene recognition accuracy can be also useful to exploit. Finally warping faces to a frontal view [36] can be useful for a more invariant face representation.